\documentclass[10pt, conference]{IEEEtran}
\IEEEoverridecommandlockouts
\usepackage[utf8]{inputenc}
\usepackage{amsmath,amssymb,amsfonts}
\usepackage{textcomp}
\usepackage[table,xcdraw]{xcolor}
\def\BibTeX{{\rm B\kern-.05em{\sc i\kern-.025em b}\kern-.08em
    T\kern-.1667em\lower.7ex\hbox{E}\kern-.125emX}}

\usepackage{algorithm}
\usepackage[noend]{algpseudocode}
\usepackage{float}
\usepackage{caption}
\usepackage{multirow}  %% For table
\usepackage{amsmath}
\usepackage{titlesec}
\titlespacing*{\section}{0pt}{1.1\baselineskip}{\baselineskip}
\usepackage{cellspace, hhline}
\usepackage{mathtools}
\usepackage{subfig}

\usepackage{breqn}
\usepackage[colorlinks=true, citecolor=blue,linkcolor=blue,bookmarks=false]{hyperref}
\usepackage{amssymb}
\usepackage[export]{adjustbox}
\usepackage[symbol]{footmisc}
\def\algbackskip{\hskip-\ALG@thistlm}

\usepackage[switch]{lineno}
\newcommand\nnfootnote[1]{%
  \begin{NoHyper}
  \renewcommand\thefootnote{}\footnote{#1}%
  \addtocounter{footnote}{-1}%
  \end{NoHyper}
}

\newcommand\githubref[1]{%
  \renewcommand\thefootnote{}\footnote{#1}%
  \addtocounter{footnote}{-1}%
}

\makeatletter % <=======================================================
\newcommand{\algmargin}{\the\ALG@thistlm}
\makeatother
\algnewcommand{\parState}[1]{\State%
    \parbox[t]{\dimexpr\linewidth-\algmargin}{\strut\hangindent=\algorithmicindent \hangafter=1 #1\strut}}
\usepackage{xcolor,soul,lipsum}
\newcommand{\myul}[2][black]{\setulcolor{#1}\ul{#2}\setulcolor{black}}

\DeclareRobustCommand*{\IEEEauthorrefmark}[1]{%
\raisebox{0pt}[0pt][0pt]{\textsuperscript{\footnotesize #1}}%
}
\usepackage[nodisplayskipstretch]{setspace}
\usepackage[style=ieee]{biblatex} % Use biblatex-iee package by importing like this 
\usepackage{caption}
\usepackage{multicol}% http://ctan.org/pkg/multicols
\usepackage{adjustbox}
\bibliography{references}
\begin{document}

\title{EvoSTS Forecasting: Evolutionary Sparse Time-Series Forecasting}

\author{\IEEEauthorblockN{Ethan Jacob Moyer\IEEEauthorrefmark{1}\IEEEauthorrefmark{*}, Alisha Isabelle Augustin \IEEEauthorrefmark{3,4}, Satvik Tripathi\IEEEauthorrefmark{2,4}, Ansh Aashish Dholakia\IEEEauthorrefmark{2}, \\ Andy Nguyen\IEEEauthorrefmark{2}, Isamu Mclean Isozaki\IEEEauthorrefmark{2}, Daniel Schwartz \IEEEauthorrefmark{2}, Edward Kim\IEEEauthorrefmark{2}}

\IEEEauthorblockA{
\IEEEauthorrefmark{1}School of Biomedical Engineering, Drexel University, PA\\
\IEEEauthorrefmark{2}College of Computing \& Informatics, Drexel University, PA\\
\IEEEauthorrefmark{3}College of Engineering, Drexel University, PA\\
\IEEEauthorrefmark{4}College of Arts and Sciences, Drexel University, PA\\
% \IEEEauthorrefmark{4}College of Arts \& Sciences, Drexel University, PA\\
Email: \{ ejm374, aia43, st3263, aad356, an839,  imi25, des338, ek826 \}@drexel.edu}}

\maketitle
\begingroup\renewcommand\thefootnote{\textsection}
\endgroup

\begin{abstract}
In this work, we highlight our novel evolutionary sparse time-series forecasting algorithm also known as EvoSTS. The algorithm attempts to evolutionary prioritize weights of Long Short-Term Memory (LSTM) Network that best minimize the reconstruction loss of a predicted signal using a learned sparse coded dictionary. In each generation of our evolutionary algorithm, a set number of children with the same initial weights are spawned. Each child undergoes a training step and adjusts their weights on the same data. Due to stochastic back-propagation, the set of children has a variety of weights with different levels of performance. The weights that best minimize the reconstruction loss with a given signal dictionary are passed to the next generation. The predictions from the best-performing weights of the first and last generation are compared. We found improvements while comparing the weights of these two generations. However, due to several confounding parameters and hyperparameter limitations, some of the weights had negligible improvements. To the best of our knowledge, this is the first attempt to use sparse coding in this way to optimize time series forecasting model weights, such as those of an LSTM network.


\end{abstract}

\nnfootnote{\IEEEauthorrefmark{*} Corresponding author}
\githubref{All source code is open-sourced at \href{https://github.com/drexelai/sparse-time-series-forecasting}{\color{blue} \myul[blue] {GitHub.}}}

\section{Introduction} \label{introductionsection}
% \IEEEPARstart{T}{he} 
Long Short-Term Memory (LSTM) networks are well-established in the field of machine learning as robust time-series forecasting models \cite{cao2019financial} \cite{sagheer2019time} . However, they often suffer from drifting predictions and compounding errors when predicting far into the future \cite{tripathi2021artificial}. Time-series modalities are ubiquitous in the field of medicine. Although this paper focuses on an Electrocardiogram (ECG) signal, our findings can be extended to other waveform, such as electroencephalogram (EEG), intracranial pressure (ICP), and mean arterial pressure (MAP). Often, multi-step predictions are far more valuable than single-step predictions as they can offer more insight into the direction of the signal. In the case of these medical signals, predicting farther into the future provides better insight into the direction of the health of a patient.  It is for this reason that advanced machine learning models and algorithms must be used in order to best predict more than just the next single time point. In this work, we attempt to improve a multi-step univariate LSTM model through a novel algorithm. 

This paper is organized such that \autoref{backgroundsection} discusses background and related work, \autoref{methodsection} discusses the implemented search algorithms, \autoref{resultssection} reports the experiments and results, \autoref{conclusionsection} concludes the paper by going over the important and discussing future work in this area of research.

\section{Background} \label{backgroundsection}

\subsection{Time Series Forecasting}

Time series data is simply a set of data points ordered by time. Time series forecasting refers to forecasting or predicting the future signal so many time points in the future. This is an important area in machine learning due to the problems often faced when making these predictions. Univariate and multivariate represent two approaches to statistical time series analysis. Univariate time series involves one variable that is varying over time. Most multivariate time series involve a dependent variable and multiple independent variables varying over time. Forecasting a multivariate time series is challenging task to due complexity, time dependency, and non-linearity. While traditional data sets can often have structural relationships expressed as a table, time series data sets require searches for behaviors and patterns in events streaming across time via specific sequences. This can be done through sampling; however, random sampling generally occurs outside the population as the time periods that are to be predicted. In addition, many domains of time series data contain non-linearities that require reliable statistical methods to summarize and understand them. As a result, the inclusion of non-linearities proves to be a complicated and difficult problem to solve \cite{potter1999nonlinear}. Generally, time series forecasting describes predicting the observation at the next time step. This is called a single-step forecast, as only one time step is to be predicted. For instance, \cite{mandal2010new} proposed a Recurrent Neural Network (RNN) to forecast the price of electricity for the next day using a single-step ahead where the networks is then applied recursively using the previous prediction as input for the subsequent forecasts. There are some multi-step time series problems where multiple time steps must be predicted \cite{kline2004methods,taieb2010multiple}. One specific example of a multi-step time series problem is forecasting the electric load using different multi-step methods, such as Auto-Regressive Integrated Moving Average (ARIMA) and a Long Short-Term-Memory (LSTM) that resulted in the LSTM model with superior performance in comparison to the ARIMA model for multi-step electric load forecasting \cite{masum2018multi}.

\subsection{Sparse Coding} \label{sparsecodingsection}

Sparse coding is a new method for finding deep patterns in unlabeled input data. In short, it learns basis functions to capture higher-level features in the given discrete data . J. Yang, Wright, Huang, and Ma (2008) were the first one to apply sparse signal representation for super resolution \cite{yang2008image-1} and since then it has been a very active area of study \cite{lu2012geometry-2, yang2010image-3, gao2012image-4}.

In natural pictures, as sparse coding is used, the learnt bases look like the receptive neuronal fields in the visual cortex \cite{hyvarinen2000emergence-5,olshausen1997sparse-6}. In contrast to other unsupervised learning approaches like principal component analysis (PCA), sparse coding could be used to learn overcomplete basis sets, where the number of bases exceeds the input dimension. Sparse coding can also sparse the activations of their models by inhibiting between bases. In biological neurons, comparable characteristics are observed, suggesting sparse coding a probable architecture of the visual cortex \cite{olshausen1997sparse-6,olshausen2004sparse-7}.

The basic notion underlying sparse signals representation, or sparse coding, is that signals can be reconstructed from their low-dimensional projections correctly by utilizing linear relationship between the signals \cite{lee2007efficient}. By reducing the complexity of input signals, sparse representations allow for faster processing and storage as well as enhanced feature extraction and pattern detection \cite{olshausen1997sparse-6, wright2010sparse-8}. The goal of sparse coding is to represent a signal of vector $x$ as a linear combination of features from a dictionary of features $D$ using a sparse set of coefficients $a$. This aims to reduce the number of features in the input signal.
\\
The goal of sparse coding may be summed up mathematically as minimizing an energy function, which is defined as-
\begin{equation}
\min _{a}\left(\left|x-D a^{T}\right|_{2}+\lambda|a|_{0}\right)
\end{equation}

To compute this minimization function, we need to optimize the dictionary reconstruction error while remaining sparsity in our dictionary. One way to do this is by using a gradient descent method. This method will give us a global optimal value since we add two convex functions to get the result. Although a gradient descent approach is appropriate, it is not differentiable for a single hidden unit. That is, the L1 norm is not differentiable at 0. Therefore, we must use a different approach that is close to gradient descent. In this, if the solution of the minimizing function changes its sign because of L1 norm gradient, we can clamp the latent representation unit to 0. So we apply these updates to all units and we repeat the procedure till the values of hidden units do not change drastically. 

Several sparse coding algorithms have been developed for sparsity techniques in MRIs, reconstruction and segmentation in medical imaging, and imaging genetics \cite{ghasemi2021afdl, lin2014sparse,huang2016advanced, tripathi2022fairness}. 

% \color{red}
% Have we seen any other attempts to use sparse coding in waveform/time series data?
% I think we should talk about this paper in this section
% $https://www.sciencedirect.com/science/article/abs/pii/S1568494618302084$
% \color{black}

\subsection{Evolutionary Algorithm}

Evolutionary algorithms stem from the field of evolutionary computing, which aims to link natural selection to computational problem solving \cite{eiben2003introduction}. Coined by Charles Darwin, natural selection is a biological process through which living organisms adapt and change to their surroundings. This concept can be extended through simulating specific processes in order to arrive at candidate solutions to a problem. Although these algorithms can have many different types, the underlying principle is more or less the same. A well-defined set of individuals in a population within an environment compete for limited resources, and the best fit individuals from the population pass on their information to a following generation. Because these problems are often stochastic in nature, the set of individuals tends to be random in any given generation of the algorithm. However, a fitness function corresponding to the limited resources in the environment selects a set of these individuals as most fit. 

Within each generation, there must be a force that forces competition amongst the individuals in the environment. Between each generation, there must be variation amongst the individuals that selects for a specified trait.

Evolutionary algorithms have become a popular means for solving difficult combinatorial optimization problems. The main proposition behind evolutionary algorithm is to iteratively improve candidate solutions over many generations. With this in mind, we generated an initial generation using a population of randomly generated weights from the original LSTM model initialization. We selected the best set of weights that result in predictions with the lowest reconstruction loss for that generation. Then, the children in the next generation used these best weights to initialize the LSTM model. In this way, each generation aims to improve its LSTM model predictions due to an advantage of having a selection of weights whose predictions best match the original sparse coded dictionary.

Evolutionary algorithms can be applied to the MNIST data-set. It is also a possible solution to the famous travelling salesman problem and is used in robotics as well. Furthermore, it has been used in various real-life applications such as code-breaking, data centers, image processing, electronic circuit design, and artificial creativity. Some previous work has explored how evolutionary algorithms can be applied to time series forecasting through Genetic Algorithms (GA), Differential Evolution algorithms (DE) and through the estimation of Distribution Algorithms (EDA) \cite{donate2013time}. This study applied these three evolutionary methods to problems relating to number of passengers of an international airline, monthly air temperature, monthly closings of the stock market, number of births in a month, as well as the comparison of differential equations.

\section{Methods} \label{methodsection}

\subsection{Data Set} \label{datasetsection}

In this paper, used the PTB diagnostic  ECG data-set from PhysioNet \cite{goldberger2000physiobank}. This data set was put together by Michael Oeff, a Professor of the department of cardiology at the Free University of Berlin. It includes 549 records from 290 patients.  The ages of the patients range from 17 to 87 years old with a mean of 57.2. Out of the 290 patients, 209 are men and 81 are women. Each patient has one to five records. Every single record includes 15 calculated signals. Of these 15 signals, there are 12 conventional ones - i, ii, iii, avr, avl, avf, v1, v2, v3, v4, v5, v6 together with the 3 Frank lead ECGs (vx, vy, vz). We digitize each signal at around 1000 samples per second, with 16-bit resolution over a range of ± 16.384 mV. The header (.hea) file in most of these ECG records represents a detailed clinical summary, stating the age and gender of the patient followed with their diagnosis, medical history, medication, interventions, coronary artery pathology, ventriculography, echocardiography, and hemodynamics. We tested our algorithm on data from the first channel from 50 of these patient files in the data set. \cite{goldberger2000physiobank}.

\begin{figure}[H]
    \centering
    \includegraphics[width=0.9\linewidth]{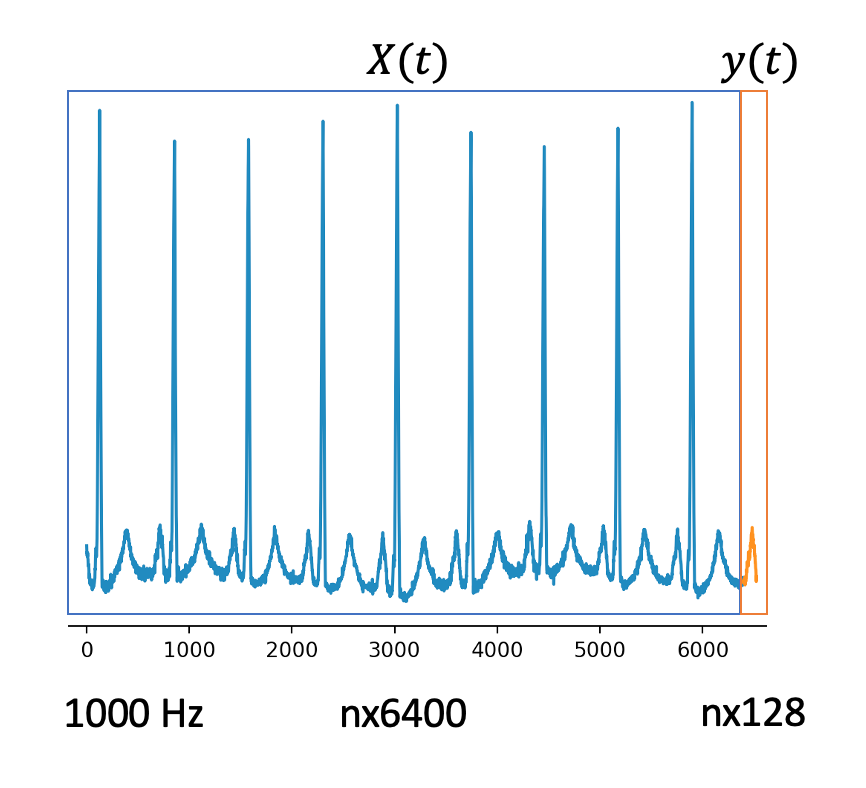}
    \caption{Example Electrocardiogram (ECG) data at 1000 Hz where the blue data points are used for features $(n=6400)$ and the orange data points are used for targets $(n=128)$.}
    \label{fig:algorithmfigure}
\end{figure}

\subsection{Network Architecture} \label{networkarchsection}

% \color{red}
% We should include a few comments about the network architecture for our a) sparse model and b) LSTM model. See commented out block below for basic table set up for the networks. Also we should explain how the ISTA model works a bit.
% \color{black}

\begin{table}[!htbp]
\centering
 \makebox[0.4\linewidth]{
 \begin{tabular}{|c|c|c|} 
 \hline
 \textbf{Layer type} & \textbf{Output Shape} & \textbf{Param} \# \\ [0.5ex]
 \hline
 LSTM  & (1, 6400)  & 2600400 \\ 
 \hline
 Dense  & (128)    & 12928 \\ 
 \hline
 \end{tabular}
 }
\caption{Architecture for LSTM Model. Discriminator has a total of 2,613,328 parameters, of which 2,613,328 are trainable and 0 are non-trainable.}
\label{tab:lstm}
\end{table}

\subsubsection{LSTM Model}
In \autoref{tab:lstm}, we have shown that the LSTM model has a total of 2,613,328 trainable parameters. The model was trained for 30 epochs with early stopping. 

\subsubsection{Sparse Model}
Our sparse model used the Iterative Shrinkage and Thresholding Algorithm (ISTA). It is a commonly used model for solving inferencing problems like dictionary learning. The algorithm initializes a vector randomly and updates it based on the reconstruction error gradient. It continues until these values converge. Then, a shrinking function is applied onto the vector to zero out units of updates would change its sign, otherwise shrink the hidden unit. 

After the ISTA model finishes, the result is a sparse representation of the input. The overall idea is that the hidden units that better explain the input are nonzero. The hidden units are compared among each other to explain the input. The lesser explained hidden units that are nonzero are overshadowed by better explained units which causes the shrink to zero. 
% \color{red}
% Is this explanation correct?
% \color{black}

% \color{orange}

% The ISTA model is known to converge if 1 over the step size is bigger than the value of the eigenvalue of a matrix.

% \color{black}

\subsection{Algorithm}
Let \textit{l} be the number of generations and \textit{k} be the number of children in a given generation.
Using a evolutionary algorithm, we obtain  optimal weights for each generation from the \textit{k} children. \cite{bas2014modified}. Each set of weights are scored based on how well their respective LSTM predictions can be reconstructed using a learned dictionary \textit{D}. This dictionary is learned through a sparse coding algorithm described in \autoref{sparsecodingsection}. The algorithm is illustrated in \autoref{fig:algorithmfigure}.

% \color{red}
% Add more about the algorithm to this section
% \color{black}

\begin{figure*}[ht]
    \centering
    \includegraphics[width=0.8\linewidth]{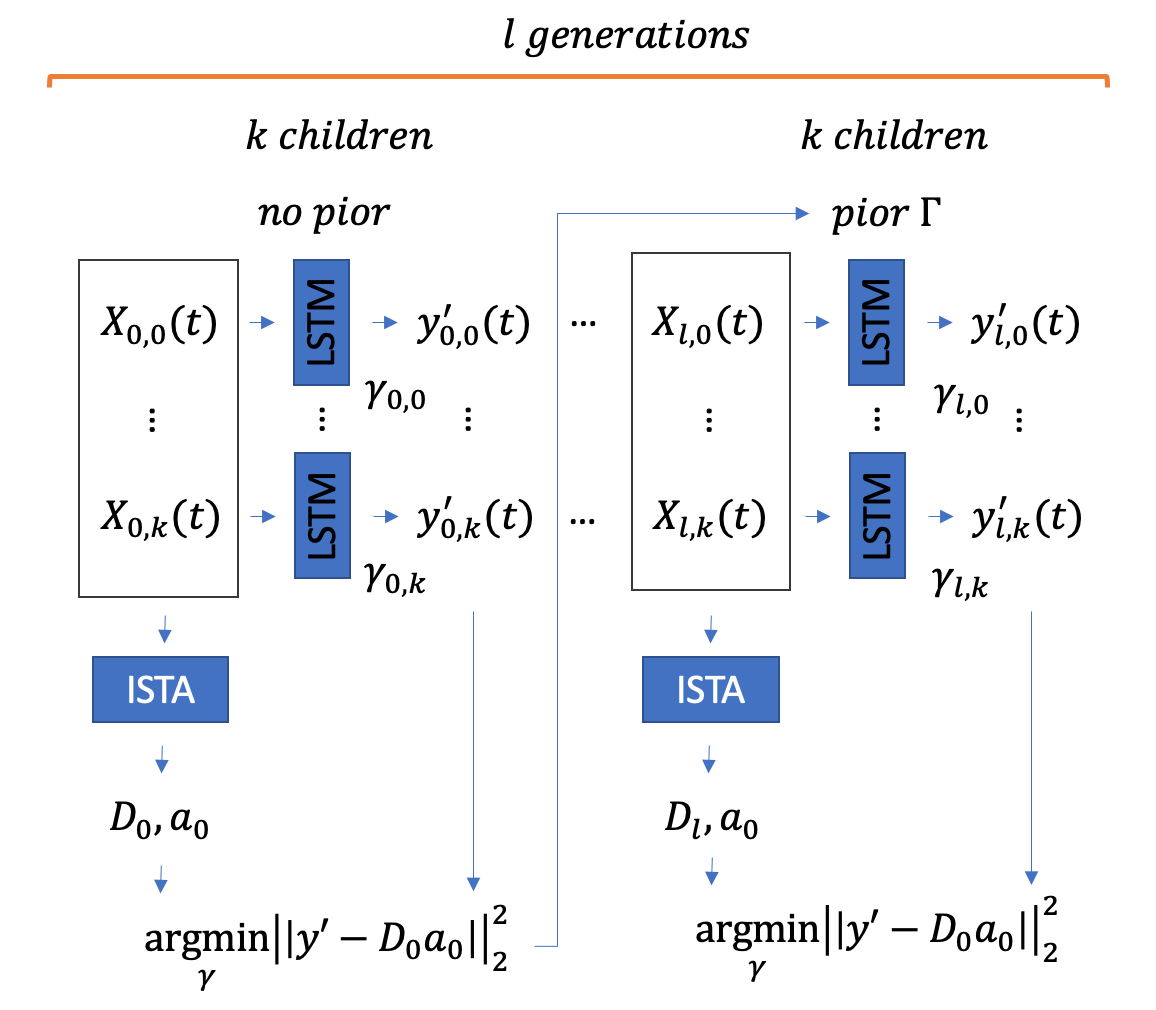}
    \caption{Evolutionary Sparse Time-Series Forecasting Algorithm (EvoSTS). Let \textit{l} be the number of generations and \textit{k} be the number of children in each generation. Each generation uses a partition of the data set \textit{X} and \textit{y}; In the Iterative Shrinkage and Thresholding Algorithm (ISTA) model, \textit{X} is used both to find a sparse dictionary representation describing the partition and to predict \textit{y}. Each child represents a set of weights $\gamma$ with an associated initialized Long Short-Term Memory (LSTM) network. In the first generation, the weights are randomly initialized. After each child in a given generation has optimized its weights and made their respective predictions for \textit{y'}, the dictionary for the partition is used to reconstruct \textit{y'}. The $\gamma$ that results in the \textit{y'} that minimizes this reconstruction is used as the prior for the next generation, designated as $Gamma$. The algorithm currently does not have any stopping criteria and continues until all generations have been evaluated.}
    \label{fig:algorithmfigure}
\end{figure*}

\section{Experiment Results \& Observations}
\label{resultssection}

To determine whether our algorithm found weights that improved the LSTM model, we compared results from the first and last generation. The first generation represents models compiled from randomized weights, while the last generation represents models compiled from weights that have historically minimized the reconstruction loss. With this in mind, we compare these two generations by reporting the root mean squared error (RMSE) and coefficient of determination (R2) of the predictions for the best performing child in each generation. In order to avoid sampling bias, we perform 10-fold cross validation and average these metrics from each fold.

\begin{table*}[ht]
\begin{tabular}{|l|l|l|l|l|}
\hline
\textbf{Partition} & \textbf{R2 Random Weights} & \textbf{R2 Optimized Weights} & \textbf{RMSE Random Weights} & \textbf{RMSE Optimized Weights} \\ \hline
1                  & -372.99                    & -374.40                       & 1.45                         & 1.45                            \\ \hline
2                  & -0.62                      & -0.63                         & 0.32                         & 0.32                            \\ \hline
3                  & -3.08                      & -3.11                         & 0.21                         & 0.21                            \\ \hline
4                  & -2.17                      & -2.18                         & 0.37                         & 0.36                            \\ \hline
5                  & -3.17                      & -3.32                         & 0.95                         & 0.94                            \\ \hline
6                  & -6.87                      & -6.98                         & 0.28                         & 0.28                            \\ \hline
7                  & -0.08                      & -0.08                         & 1.15                         & 1.15                            \\ \hline
8                  & -0.42                      & -0.42                         & 0.70                         & 0.70                            \\ \hline
9                  & -0.06                      & -0.06                         & 1.20                         & 1.20                            \\ \hline
10                 & -15.52                     & -15.40                        & 1.41                         & 1.41                            \\ \hline
\end{tabular}
\caption{10-fold cross validation experiment to validate the results of the model.}
\label{table:results}
\end{table*}

\section{Conclusion and Future Work} \label{conclusionsection}

The limitations of this algorithm include biases related to the sparse coding algorithm given its dependant nature on repetitive signals. If there were to be an artifact or drastic change within the target time-series signal, the algorithm would have a difficult time predicting it. This is because it relies on a sparse representation of the input signal, which may not have dictionary elements to explain these artifacts. A second limitation is that the use of this algorithm is restricted to periodic signals since it relies on time-series patterns in the data. This paper focused on an ECG data set but future studies can expand upon different signals.

% \color{red}
% Mention GRU (https://datascience.stackexchange.com/questions/14581/when-to-use-gru-over-lstm), other time-series specific models. Alisha?
% \color{black}

\section*{Acknowledgment}

We would like to acknowledge Drexel Society of Artificial Intelligence for its contributions and support for this research. 

\printbibliography
I\end{document}